\documentclass[sigconf]{acmart}
\usepackage{graphicx}
\usepackage{svg}

\usepackage{tablefootnote}
\usepackage{subfig}
\usepackage{multirow}
\AtBeginDocument{%
  }

\copyrightyear{2024}
\acmYear{2024}
\setcopyright{rightsretained}
\acmConference[CODS-COMAD Dec '24]{8th International Confernce on Data Science and Management of Data (12th ACM IKDD CODS and 30th COMAD)}{December 18--21, 2024}{Jodhpur, India}
\acmBooktitle{8th International Confernce on Data Science and Management of Data (12th ACM IKDD CODS and 30th COMAD) (CODS-COMAD Dec '24), December 18--21, 2024, Jodhpur, India}
\acmPrice{}
\acmDOI{10.1145/3703323.3703332}
\acmISBN{979-8-4007-1124-4/24/12}
\newcommand{\ir}{{\tt ReFrame}} 

\begin{document}

\title{\ir: Rectification Framework for Image Explaining Architectures} 

\author{Debjyoti Das Adhikary}
\email{debjyoti.das.adhikary@kgpian.iitkgp.ac.in}
 \affiliation{%
  \institution{Indian Institute of Technology, Kharagpur}
  \city{Kharagpur}
  \state{West Bengal}
  \country{India}
}

\author{Aritra Hazra}
\email{aritrah@cse.iitkgp.ac.in}
\affiliation{%
  \institution{Indian Institute of Technology, Kharagpur}
  \city{Kharagpur}
  \state{West Bengal}
  \country{India}
}
\author{Partha Pratim Chakrabarti}
\email{ppchak@cse.iitkgp.ac.in}
\affiliation{%
  \institution{Indian Institute of Technology, Kharagpur}
  \city{Kharagpur}
  \state{West Bengal}
  \country{India}
}


\begin{abstract}
Image explanation has been one of the key research interests in the Deep Learning field. Throughout the years, several approaches have been adopted to explain an input image fed by the user. From detecting an object in a given image to explaining it in human understandable sentence, to having a conversation describing the image, this problem has seen an immense change throughout the years, However, the existing works have been often found to (a) hallucinate objects that do not exist in the image and/or (b) lack identifying the complete set of objects present in the image. In this paper, we propose a novel approach to mitigate these drawbacks of inconsistency and incompleteness of the objects recognized during the image explanation.  To enable this, we propose an interpretable framework that can be plugged atop diverse image explaining frameworks including Image Captioning, Visual Question Answering (VQA) and Prompt-based AI using LLMs, thereby enhancing their explanation capabilities by rectifying the incorrect or missing objects. We further measure the efficacy of the rectified explanations generated through our proposed approaches leveraging object based precision metrics, and showcase the improvements in the inconsistency and completeness of image explanations. Quantitatively, the proposed framework is able to improve the explanations over the baseline architectures of Image Captioning (improving the completeness by 81.81\% and inconsistency by  37.10\%), Visual Question Answering(average of 9.6\% and  37.10\% in completeness and inconsistency respectively) and Prompt-based AI model (0.01\% and  5.2\% for completeness and inconsistency respectively) surpassing the current state-of-the-art by a substantial margin. 
\end{abstract}




\maketitle

\section{Introduction}
Deep Neural Networks have surfaced as a transformative force, redefining the frontiers of technology and research. It has achieved advances across a lot of sectors, such as health care, agriculture, finance, education, manufacturing, and security, by allowing machines to recognize complex patterns and execute tasks with human-like accuracy~\cite{zhao2019deep,dwivedi2023so}. The remarkable achievements of Deep Learning (DL) have not only solved long-standing computational quandaries but have also opened pathways to new possibilities, catalyzing innovations that were once beyond imagination. The success recipe of DL applications dealing with images, texts, audio and video is attributed over the years from its ever-increasing complicacy and variations in the neural network architectures. Such prolific growth in deep network structures for solving complex problems has significantly diminished the potential of outcomes to be reasoned and scrutinized. 

One of the most fundamental problems that Deep Learning researchers deal with is the explanation of images~\cite{ngiam2011multimodal}, which often becomes a multi-modal domain due to the incorporation of textual information to guide the predicting process. Converging the field of Computer Vision and Natural Language Processing, the multi-modal image explanation focuses on explaining an input image in a human-understandable way.
Several approaches has been proposed to explain images. In particular, {\em Image Captioning}, the first of these, involves creating descriptive text for an image, encapsulating its contents in a concise sentence~\cite{hossain2019comprehensive}. Alternatively, explanation is also generated through a dynamic interaction, called {\em Visual Question Answering (VQA)}, where the system generates responses to specific questions posed about an image, showcasing a more focused understanding~\cite{antol2015vqa}. Extending this method, in the past few years, Large Language Models (LLMs) have been proposed which especially deals with the paradigm of {\em conversational or prompt-based agents}. Presently, LLMs such as ChatGPT represents the pinnacle of this domain, where the system engages in a dialogue about an image, crafting responses within a conversational context and demonstrating a deeper, more nuanced grasp of the visual stimuli~\cite{achiam2023gpt}. 

\begin{figure}[h] 
    \centering
    \subfloat{
        \centering
        \includegraphics[width=0.4\textwidth]{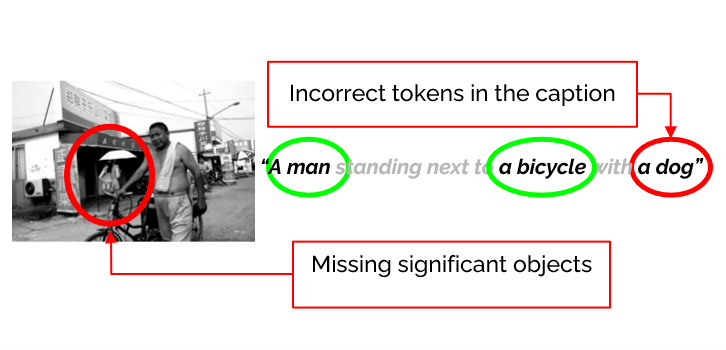}
        \label{IC_DRAWBACKS}

    }
    \\
    \subfloat{
        \includegraphics[width=0.4\textwidth]{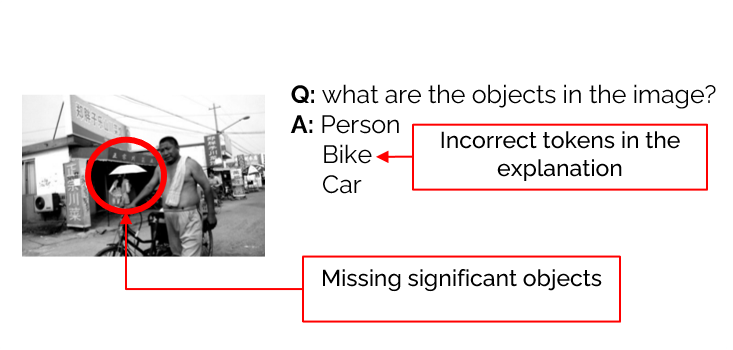}
        \label{VQA_DRAWBACKS}
    }
    \caption{Shortcomings of Baseline Multi-modal Architectures}
    \label{DRAWBACKS}
\end{figure}

Despite such advancements in generating vivid explanations from images, there remains significant gaps, primarily due to the fact that the explanations generated from can not be verified in an unsupervised environment. There are no standard process of (a) validating an auto-generated explanation, or (b) assessing the comprehensiveness of the resulting explanation. The existing works have been found to hallucinate or misinterpret objects that do not exist in the image and/or are unable to identify the complete set of objects present in the image (illustrated in Figure~\ref{DRAWBACKS} where one can see that the models are not able to recognize the person behind, and is incorrectly hallucinating the presence of a dog) ~\cite{bai2024hallucination,huang2024visual}. Hence, there is a dire need to develop image explaining models that are much more {\em consistent} and {\em complete}. Consistency can be referred to as the coherence and accuracy of the explanation with respect to the image, while completeness can be measured as the extent to which the explanation captures all relevant content. Traditional evaluation metrics such as BLEU or CIDEr score are not equipped to fully capture these aspects, often leading to an overestimation of a model’s performance when explanations are generated beyond the scope of the reference data. This necessitates us to propose a new framework along with model evaluation models to determine and attain consistent as well as complete image explanations. The work presented in this article is an enabler of the same. 

\begin{figure}[H]
    \centering
    \includegraphics[width=0.7\linewidth]{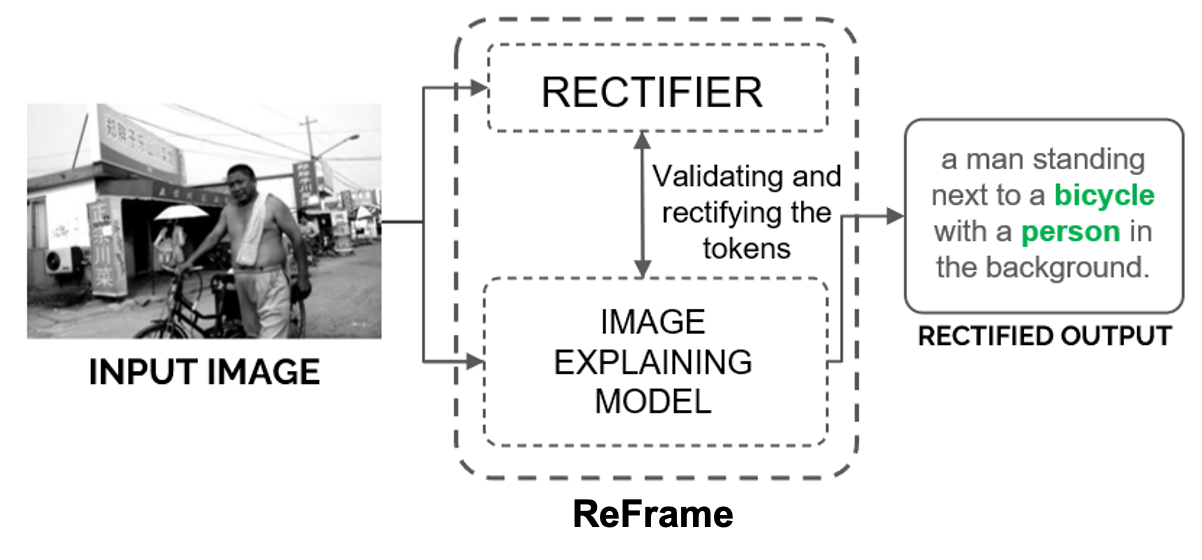}
    \caption{Improved Explainability through our Proposed Framework \ir.}
    \label{REFRAME}
\end{figure}

We introduce a novel framework, named \ir, which can be plugged on top of the existing architecture and enhances the latter's explaining capabilities. It integrates a rectification layer to refine the outputs of existing image explanation methods. This framework leverages the object detection strength of Mask R-CNN (MRCNN) model~\cite{he2017mask} and uses it as a rectifier to assess and correct the outcomes of any generic explainable architecture (such as captioning, visual question answering, or prompt-based), ensuring the generated explanations are both comprehensive and aligned with the input image. The architecture gives an interpretable rectification by providing the reasoning behind the process. Figure~\ref{REFRAME} illustrates, through the same instance in Figure~\ref{DRAWBACKS}, how the rectification step of our \ir \: framework can improve the shortcomings found in existing image explainability models followed by the architecture's performance over some sample instances.

\begin{figure*}[h]
    \centering
        \includegraphics[width=0.7\textwidth]{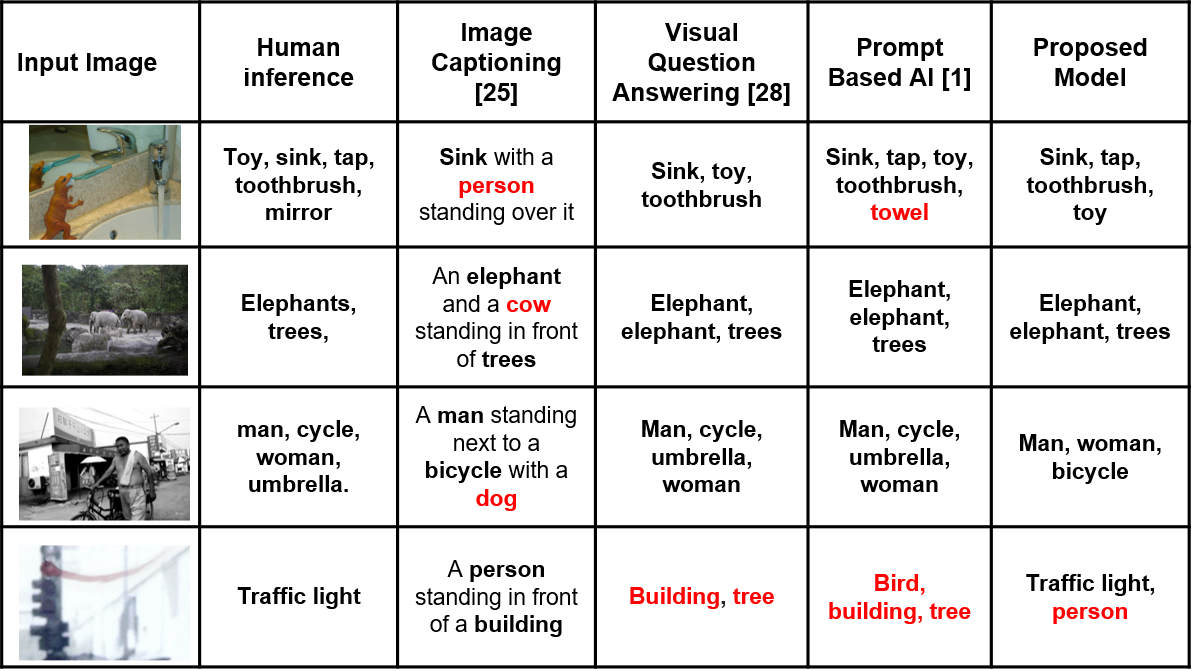}
    \caption{Instances explaining the architecture and its improvement over baseline explanations. The table illustrates the rectified objects across the 3 domains alongwith the reference of human recognized objects.}
    \label{SAMPLE_RESULTS}
\end{figure*}

Subsequently, to quantitatively assess the quality of our framework, we design evaluation metrics to provide an interpretable assessment of image explanation, particularly in unsupervised settings where models generate explanations without reliance on predefined ground truths. Since the rectified predictions made by the models are unsupervised, they cannot be compared with the ground truth, and hence the proposed metrics compute the completeness of the explanation and how consistent it is compared to the supervisor.

The primary contributions of our work can be outlined as follows:
\begin{itemize}
    \item We propose a generic framework (\ir) which can be extended on top of any existing image-explaining architecture for precise and comprehensive explanation of images.
    \item We present an unsupervised rectifier model on top of an existing framework that rectifies the explanations generated by the model, with respect to a supervisor.
    \item We devise precision metrics to evaluate the completeness and inconsistency of the explanations based on detected objects.
    \item We demonstrate the efficacy in generating image explanations by our proposed framework through improved consistency and better completeness over the state-of-the-art explanation models.
\end{itemize}

Subsequent sections of this paper are structured as follows. Section~\ref{sec:literature} exposes the current state-of-the-art methods in multi-modal deep learning for image explanations. Section~\ref{sec:framework} introduces our proposed framework (\ir) and its adoption over three different image explanability frameworks (i.e. captioning, VQA and prompt-directed). Section~\ref{sec:implementation} presents the implementation details along with the proposed evaluation metrics for the improved architecture. Section~\ref{sec:result} illustrates the experimental results and showcases the improvements. Finally, Section~\ref{sec:conclusion} concludes the discussion presented in this article. 

\section{Related Works} \label{sec:literature}
The quest for interpretable architectures in the field of image explanation has led to significant advancements in AI research. These endeavors aim to bridge the gap between complex visual data and human-like understanding, employing various approaches such as Image Captioning, Visual Question Answering (VQA), and integration with advanced language models like GPT-4. However, despite this progress, these architectures exhibit notable limitations, particularly in their interpretability, completeness, and consistency. In this section, we discuss our review of the current landscape, highlighting the inherent drawbacks of existing models and underscoring the paucity of research focused on enhancing the completeness and consistency of image explanations.

\subsection{Image Captioning}
Image Captioning is one of the widely used approaches for explaining the visual content of an image in natural language. This domain has recently undergone tremendous growth with the help of Deep Learning models. The early dawn of this domain witnessed a huge focus on the description of retrieval-based~\cite{gong2014improving}, or template-based architectures~\cite{farhadi2010every,vinyals2016show}, and novel approaches~\cite{kiros2014multimodal}. One of the biggest innovations in this field was~\cite{vinyals2015show}, which changed the field altogether.~\cite{vinyals2015show} is a very simple yet powerful approach that uses the inherent architecture of any image captioning model of having an encoder and decoder. However, the use of Convolutional Neural Network (CNN) to grasp the visual embeddings from the image and the use of Recurrent Neural Network for the textual decoder for generating the caption makes this model applicable in a lot of works till date~\cite{ge2024show,nguyen2023show,lindh2023show}. In recent years, much more sophisticated approaches have been taken up that involves use of Transformers, Attention Networks,~\cite{stefanini2022show}. One such recent novel approach is~\cite{li2022mplug} which is a transformer-based model that combines cross-modal understanding and generation and has achieved state-of-the-art results on over various datasets. In our study, we have used~\cite{vinyals2015show} and implemented our add-on rectification architecture over it, and compared it with~\cite{li2022mplug} to show the efficacy of the proposed approach. 

\subsection{Visual Question Answering}
Visual Question Answering (VQA) systems aim to interpret and respond to natural language questions about images. The domain of VQA also started with combining the strength of CNN for extracting the visual features, using the LSTM for getting the textual embeddings from the questions, and then finally predicting the result using a decoder such as LSTM or simply a fully connected layer. One of the very first groundbreaking architectures in this domain was~\cite{antol2015vqa}, which used this architecture and created a new baseline for this field. Numerous work have shown in various aspects in this domain, such as enhancing the interpretability using attention networks~\cite{yu2019deep}. In recent years, many domains involving multi-modal tasks have been observed to use Vision-and-Language Pre-training (VLP) models. Which are pre-trained using image text matching and masked language modeling techniques~\cite{li2022mplug,pan2022contrastive}. With the~\cite{kim2021vilt} being one such architecture in the field of VQA, it has shown significantly great results, achieving state-of-the-art over various datasets. In our work, we use~\cite{antol2015vqa} as our baseline and build our rectifying architecture on top of it, and ultimately, compare it with~\cite{kim2021vilt}, the current state-of-the-art model in this field to demonstrate how our proposed framework is surpassing the latter.
 
 \subsection{Conversational or Prompt-based Agents}
 In the past year, a new type of domain came into the picture and has changed the view of multi-modal processing, viz., Large Language Models. LLMs are models based on a transformer architecture, that use generative models to perform natural language processing (NLP) tasks~\cite{hu2023promptcap,shen2024hugginggpt}. This has revolutionized the field by demonstrating the power of large-scale language models that are also capable of engaging with different modalities, such as images, audio, and structured data. At the time of this work, one of the best and most renowned LLM architecture is OpenAI's Chat-GPT, a Prompt-based agent, which gives a chat interface to interact with the model~\cite{achiam2023gpt}. GPT-4, the latest iteration, has enhananced capabilities to understand and generate responses with greater contextual relevance and nuanced understanding, even in complex dialogues. This not only makes it more practical for widespread use across various platforms but also sets a precedent for future developments in AI efficiency. Hence for our, final module, we implement our MRCNN on top of the outputs generated by GPT-4.

In spite of all the advances across all the domains, we noticed that there has been a huge gap when it comes to the reasoning and rectification of image-explaining models. We observed that the focus has predominantly been on optimizing the accuracy and fluency of outputs, with little consideration for the reasoning and interpretability of the outputs by these, which can be further validated by other studies as well~\cite{rudroff2024revealing,chu2024task,wang2024exploring}. The generic descriptions provided by such models lack depth and fail to capture unique or subtle nuances present in the images. Moreover, the reliance on datasets with predefined captions for training further exacerbates this issue, as the models tend to reproduce common phrases and descriptions, neglecting the diversity and specificity required for accurate and meaningful image explanation. In an attempt to bridge this gap of reasoning and rectification, our work proposes a holistic approach over the three most renowned multi-modal domains in the current Deep Learning field that significantly improves their explaining capabilities.
\section{\ir: Our Proposed Framework} \label{sec:framework}
\begin{figure*}[htb]
    \centering
    \includegraphics[width=0.6\textwidth]{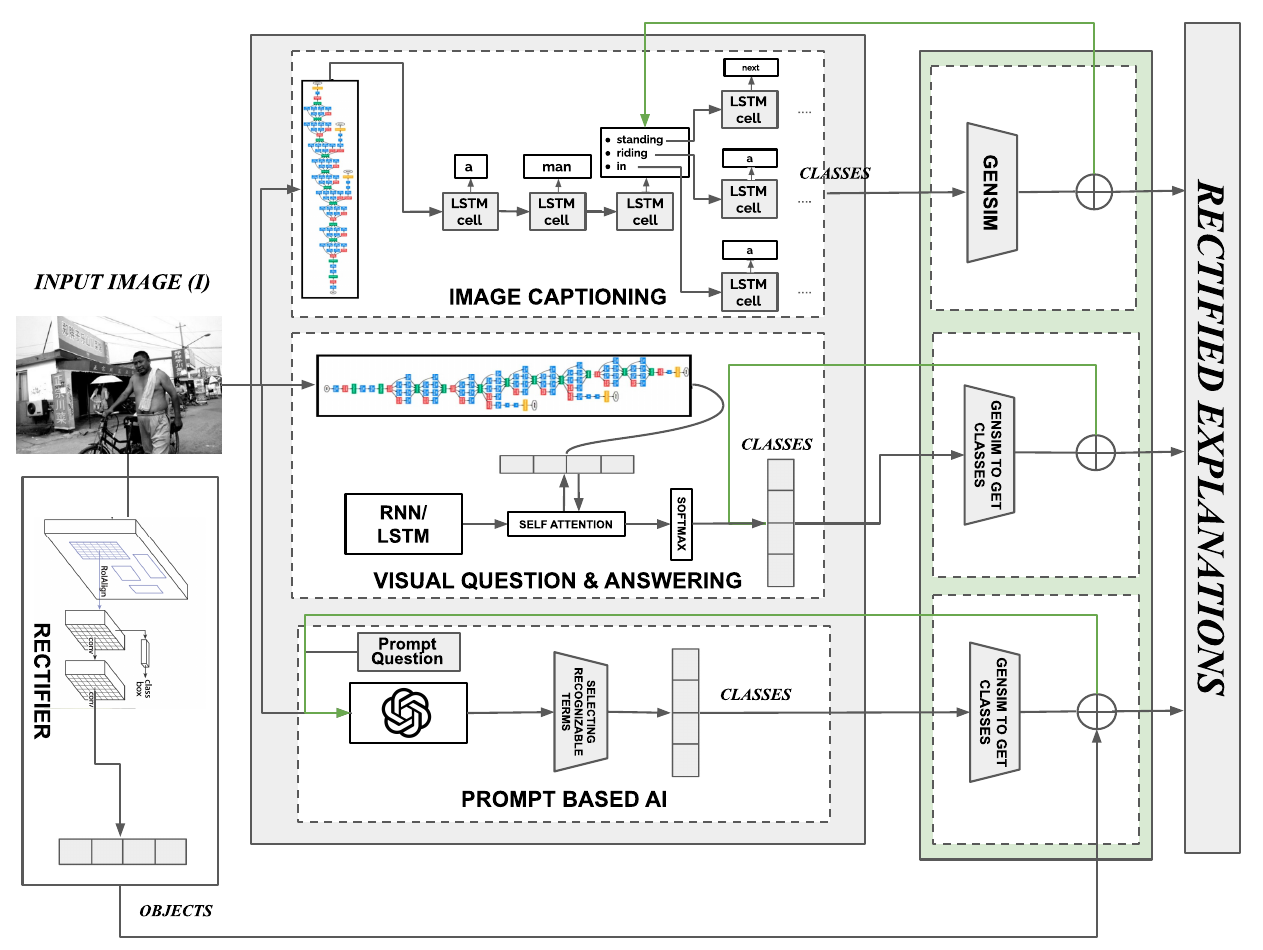}
    \caption{Architecture of our proposed model, \ir}
    \label{ARCHITECTURE}
\end{figure*}

The crux of our proposed methodology lies in the integration of a model-agnostic rectifier whose task is to enhance the accuracy and depth of image explanations irrespective of the underlying domain. This rectifier block, built by Mask R-CNN (MRCNN),  as illustrated in Figure~\ref{ARCHITECTURE}, serves as a universal solution, interfacing seamlessly with each domain to ensure consistency and completeness in the outputs. It is designed by connecting the final output layer of the underlying domains to it and then cross-verifying the objects predicted by the image explaining model to that of the rectifier.

By incorporating MRCNN as the rectification layer, the proposed system leverages its advanced capabilities to critically check and rectify the outputs generated by the underlying Image Captioning, VQA, and Prompt-based AI modules. However, this leads to two major challenges: (a) The vocabulary of the decoder's length is much higher than what can be recognized by the MRCNN, and (b) finding the correct token to be replaced with the inconsistent one. In order to overcome the first challenge we use the pretrained word2vec model of Google-News-vectors-negative-300 to map the entire vocabulary to the 80 classes that can be recognized by the rectifier~\cite{mikolov2013efficient,mikolov2013distributed}.

\subsection{Rectification over Image Captioning Models}
\begin{figure*}[htb]
    \centering
    \subfloat{
    \centering
        \includegraphics[width=0.4\textwidth]{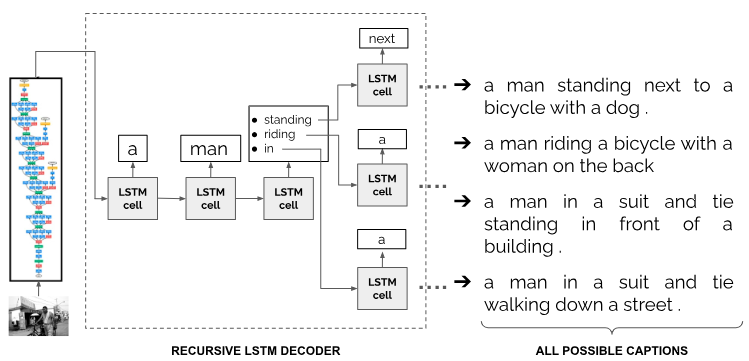}
        \label{IC_CAPTION_TREE_LSTM}   
    }
    \subfloat{
        \includegraphics[width=0.45\textwidth]{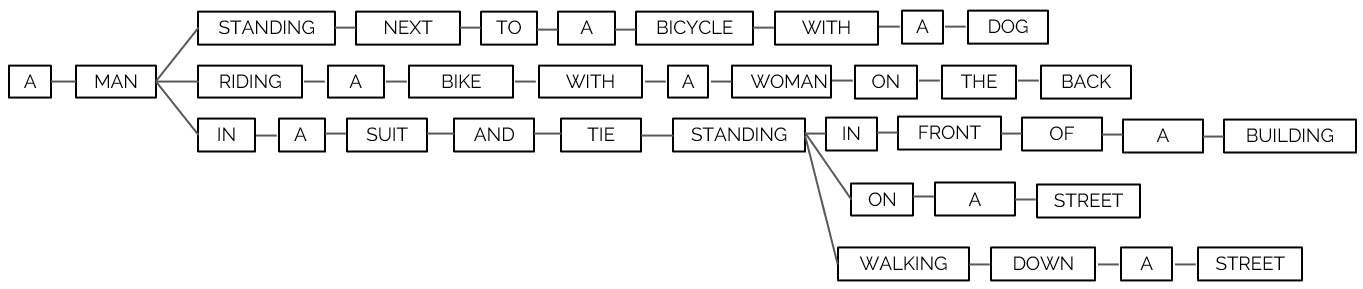}
        \label{IC_CAPTION_TREE}
    }
    \caption{Caption Tree Generation over baseline Image Captioning ~\cite{vinyals2015show}}
    \label{CAPTION_TREE}
\end{figure*}

In the domain of Image Captioning, we used Show And Tell, a highly acclaimed architecture in the field of Deep Learning~\cite{vinyals2015show}. In~\cite{vinyals2015show}, the authors proposed a methodology that extracts the visual features from an encoder, mainly a pre-trained CNN like Resnet152, VGGNet, etc., followed by a textual Decoder, which is responsible for generating the entire caption. This approach generates a single caption by considering the most probable token at each time stamp and in the process, a lot of objects that indeed exist in the image gets removed. In our approach, we consider multiple options of tokens at every time stamp and feed them back to LSTM in a recursive manner, hence generating an essence of branches of captions as illustrated in Figure~\ref{CAPTION_TREE}.

This caption tree was generated by redefining the LSTM in a recursive manner, where, instead of one hidden state vector, we send a set of vectors back to the recurrent unit with a recursive approach so that each call gets a different memory stack and hence getting the common past context and simultaneously generating different captions from different hidden states. This new recursive approach to LSTM, allows us to generate multiple relevant captions.

Once we have this set of captions with us, we use our rectifier on top of this captioning model for every timestamp of all the branches. Whenever the explanation generates a potential object, we check the existence of that object in the set of classes that have been recognized by the rectifier. The rectifier then feeds the token with the highest probability of the decoder that exists in the objects recognized by MRCNN as well. This helps mitigate the problem of inconsistent objects being recognized by the baseline decoder. These generated set of captions turns to have much more ingrained explanations with a lesser inconsistency.

\subsection{Rectification over Visual Question Answering Models}
\begin{figure}[htb]
    \centering
    \includegraphics[width=0.6\linewidth]{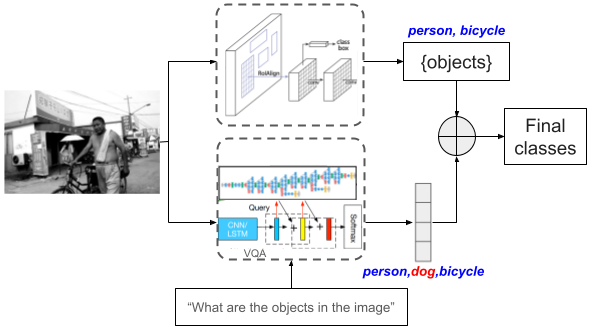}
    \caption{Rectification over baseline VQA model.}
    \label{VQA_RECTIFICATION}
\end{figure}

For the module of Visual Question Answering, we used~\cite{yang2016stacked} as our baseline since it was the very first innovative approach to encounter this problem. As illustrated in Figure~\ref{VQA_RECTIFICATION}, we adopt a similar approach to~\cite{vinyals2015show} using a pre-trained CNN to extract the visual features. In order to learn the question embedding, the authors have used an additional textual encoder. The visual embeddings and the questions embedding vectors are then concatenated to be fed to a fully connected layer which then generates the final answer.

Similar to the domain of Image Captioning, we noticed that VQA also follows the same pattern of giving incorrect explanation for the question, with the correct explanation being slightly down the probability score of the full connected layer. The rectifier then checks, validates and rectifies the results predicted in the final output layer. We considered the topk (k {$\in$ \{5,10,15\}}) outputs from the output layer of the VQA model and then check the answer with the highest probability that is agreed upon by both the base model and the rectifier. This approach helps double-check the final answer being predicted by the enhanced VQA model.

\subsection{Rectification over Prompt-based AI Model}
For our last domain in multi-modal image explaining domains, we experimented with a well-known model of Large Language Model, the Generative Pre-trained Transformer series by OpenAI, GPT-4~\cite{achiam2023gpt}. In our proposed architecture \ir, we innovatively repurposed GPT-4, enabling it to not only generate explanations for images but also to engage in a self-rectification process based on feedback from the Rectifier as shown in the main Figure~\ref{ARCHITECTURE}. Initially, GPT-4 is prompted to describe the objects in an input image to produce initial explanations using its vast language and image processing capabilities. These descriptions are then cross-referenced against the rectifier to identify any inconsistencies or omissions in GPT-4’s output. Using the classes fed by the rectifier, GPT-4 receives a re-prompt to incorporate the former's findings. This mechanism enables GPT-4 to rectify upon its initial output, making it more aligned with the objects recognized by the Rectifier. This iterative rectification process serves to refine the explanations, leveraging the strengths of GPT-4’s generative capabilities and MRCNN’s precise object detection to produce more consistent and complete image explanations.
\section{Implementation Details and Evaluation Metrics} \label{sec:implementation}
\subsection{Dataset Description}
In our work, since we are targeting to build one single framework to be implemented across multiple multi-domain architectures, the dataset employed plays a crucial role. Both, Image Captioning and Visual Question Answering models were pre-trained on the MSCOCO training set. The dataset details is as follows:
\begin{itemize}

    \item Comprises 118,287 images, each annotated with 5 different explanations.
    \item Designed specifically for object recognition, segmentation, and captioning tasks~\cite{lin2014microsoft}.
    \item Provides a rich and varied linguistic context for each image.
\end{itemize}

For the Prompt-based AI module, we utilized GPT-4, which, although not trained specifically on the MSCOCO dataset, benefits from expansive training on diverse datasets. The Mask R-CNN architecture employed as the rectifier, utilizing a ResNet-FPN-50 backbone, was also pre-trained on the MSCOCO dataset. This helped ensure consistency across all the 3 domains to be rectified. The linguistic component of our model utilizes the gensim model `GoogleNews-vectors-negative300.bin', a pre-trained word vector model developed by Google. This model consists of of 300-dimensional vectors for 3 million words and phrases, trained on part of the Google News dataset.
The evaluation was done as follows:
\begin{itemize}
    \item Utilized MSCOCO validation set, containing 5,000 images.
    \item Serves as a testing ground for proposed metrics of `inconsistency' and `completeness'.
    \item Applied metrics to measure the performance of models in accurately and comprehensively describing depicted objects.
\end{itemize}

\subsection{Implementation Setup}

\subsubsection{Image Captioning Implementation Setup}
The Image Captioning module of our study employs~\cite{vinyals2015show}, which consists two major components: an encoder and a decoder. The encoder used for our experiment is ResNet-152, a Convolutional Neural Network known for its efficacy in extracting high-level features from images ~\cite{he2016deep}. This choice was motivated by ResNet-152's robust performance in Image Classification tasks, making it an ideal candidate for providing rich visual features for caption generation. The decoder component of the Image Captioning model is built using a Long Short-Term Memory (LSTM) network, which is adept at handling sequential data and generating coherent textual descriptions. Both the embedding size and the size of the hidden layers within the LSTM decoder are set to 512, optimizing the model's capacity to synthesize detailed and contextually relevant captions from the encoded image features. For the process of rectification, the decoder's final output layer is subjected to an additional verification step by the rectifier. Finally, as a benchmark for state-of-the-art performance, we use the pre-trained model of ~\cite{li2022mplug} to experiment over it and compute the inconsistency and completeness of the model for Image Captioning, over the same dataset to compare the results with that of our proposed approach.

\subsubsection{Visual Question Answering (VQA) Implementation Setup}
For the VQA component, our baseline model employs the Stacked Attention Network (SAN), known for its effectiveness in handling complex visual and textual queries ~\cite{antol2015vqa}. The model uses VGGNet as an image encoder and LSTM as the encoder for generating the final output for which it uses a Fully connected layer of length equal to the vocabulary of the dataset. The SAN model iteratively applies attention mechanisms to focus on relevant parts of the image in response to a given question, facilitating a nuanced understanding of the visual content. The VQA model is provided with the prompt of "What are the objects in the image?" and lists topk recognizable objects from the image. In order to make the observation thorough, we chose 3 valuse for k={5,10,15}. These listed objects are then checked by the MRCNN for their existence in the image. If the VQA gives objects that is not recognized by the MRCNN, the object is replaced with the next most probable object that is agreed upon by both, hence assuring consistency. Once implemented, we compute the completeness and inconsistency scores of the baseline and the baseline along with the rectifier on top of it. We then experiment over ~\cite{kim2021vilt} and compute the inconsistency and completeness scores of the model over the same dataset.

\subsubsection{Prompt-based AI with GPT-4 Implementation Setup}
In the domain of Prompt-based AI, we employ GPT-4, the latest iteration of OpenAI's Generative Pre-trained Transformer series. Due to the proprietary nature of GPT-4, direct integration with the rectification layer was not feasible. In order to simulate the same experimental environments in this domain, we manuallay asked the model by feeding the prompt "What are the objects in the image?". Following the outputs of the model, the rectification prompts by the outputs of MRCNN, were manually fed into the GPT-4 model. By mentioning the "<inconsistent objects> does not exist in the image", and then asked the same question again. This approach allows us to simulate the rectification process, enhancing GPT-4's responses in image-based conversational scenarios by ensuring that the model's outputs are consistent with the detected visual elements and their attributes.

\subsection{Metrics for Evaluation}
Now that we have an enhanced set of rectified explanations, we need to quantify the extent of their comprehensiveness and their agreement with the rectifier.  Since our rectification algorithm is unsupervised and takes the caption beyond the ground truth, the existing metrics (BLUE,CIDER,SPICE,etc.,) would not be able to gauge the efficiency of our algorithm in a just manner as they compare the given text to the ground truth. To address this, we propose two novel metrics: {\em Inconsistency} and {\em Completeness}, designed to measure the extent to which image explanations align with or diverge from the visual realities as identified by the rectifier.

\subsubsection{Inconsistency:} The Inconsistency metric evaluates the precision of image explanations by measuring the error rate in terms of falsely identified objects. As illustrated in Figure~\ref{METRICS}, the inconsistency score is calculated by identifying the instances where the image explaining model (e.g., an Image Captioning or VQA model) asserts the presence of objects and the rectifier does not detect in the image. This is computed by getting the set difference between them. The frequency of these instances is then normalized by the total number of objects identified by the rectifier to provide a ratio representing the inconsistency of the explanations with the actual image content.
\begin{equation}
Inconsistency =   \frac{\# \big[ \{Detected~Objects\} - \{Objects~with~Rectification\} \big]}{\#\big[ Objects~with~Rectification \big]}
\end{equation}
\subsubsection{Completeness:} 
\begin{figure*}[h] 
    \centering
    \subfloat{
        \centering
        \includegraphics[width=0.2\textwidth]{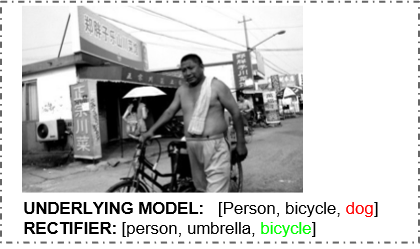}
        \label{INPUT_IMAGE}
    }
    \subfloat{
        \includegraphics[width=0.2\textwidth]{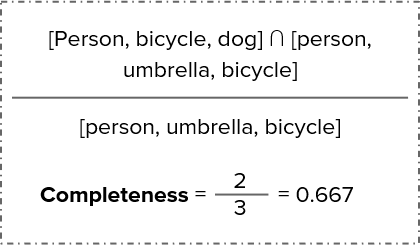}
        \label{COMPLETENESS}
    }
        \subfloat{
        \includegraphics[width=0.2\textwidth]{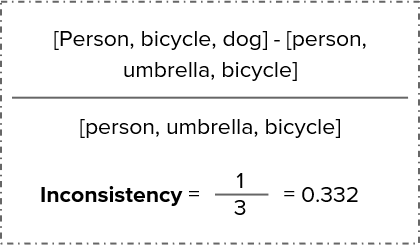}
        \label{INCONSISTENCY}
    }
    \caption{Inconsistency and Completeness over sample Input Image. The image in left represents the classes identified by any of the image explaining models, and the the classes recognized by the rectifier in the input image.}
    \label{METRICS}
\end{figure*}

The Completeness metric indicates the model's ability to provide comprehensive and accurate depictions of visual content, reflecting its effectiveness in capturing the full scope of the image. To calculate the completeness score, we tally the instances where the image explaining model accurately identifies and describes objects that are also detected and confirmed by the rectifier as illustrated in Figure~\ref{METRICS}. This count is then normalized by the total number of objects identified by the rectifier, offering a ratio that represents the completeness of the image explanation in capturing and accurately reflecting the visual content.
\begin{equation}
    Completeness =   \frac{\#\big[ \{Detected~Objects\} \cap \{Objects~with~Rectification\} \big]}{\#\big[ Objects~with~Rectification \big]}
\end{equation}

The development of image-explaining models has traditionally relied on conventional metrics such as the BLEU score to evaluate the accuracy and comprehensiveness of generated textual descriptions. However, these metrics are inherently limited as they compare generated explanations directly against a labeled ground truth. Since. our proposed model's rectification goes beyond the provided ground truth, we cannot use the conventional metrics for evaluating its performance.

\section{Experimental Results and Discussion} \label{sec:result}

Our proposed architecture presents a novel interpretable approach for the rectification of image-explaining models, which integrates Masked R-CNN (MRCNN) as a rectification layer atop the baseline models. Implementing the proposed model over the dataset, the results demonstrate substantial improvements across all evaluated domains, positioning our architecture as a significant advancement over state-of-the-art models, including GPT-4 in the context of Prompt Based Agents.

\begin{table}[h]
\centering
\scriptsize
\caption{Performance of our Proposed Approach for Image Captioning}\label{tab1}
\begin{tabular}{|l||c|c|c|}
\hline
                                                                                            & {\bf Inconsistency} & {\bf Completeness}  & {\bf BLEU Score}     \\ \hline \hline
Ground Truth (validation set)                                                               & 0.304         & 0.348         & NA             \\ \hline
Show And Tell ~\cite{vinyals2015show}                                                                              & 0.318         & 0.297         & \textbf{0.606} \\ \hline
\begin{tabular}[c]{@{}l@{}}Show and Tell + Rectification\\ (Proposed Approach)\end{tabular} & \textbf{0.200} & \textbf{0.540} & 0.596          \\ \hline
\end{tabular}
\end{table}

\begin{table}[h]
	\centering
	\caption{Completeness Results across Different Image Explaining Models}
	\footnotesize
	\label{tab3}
	\begin{tabular}{|l||c|ccc|c|}
		\hline
		\multirow{2}{*}{}                                                            &
		\multirow{2}{*}{\begin{tabular}[c]{@{}c@{}}Captioning \\ Model\footnotemark[1]\end{tabular}} &
		\multicolumn{3}{c|}{VQA Model\footnotemark[2]}                                              &
		\multirow{2}{*}{\begin{tabular}[c]{@{}c@{}}Prompt-based \\ Model {[}1{]}\end{tabular}}                                                                           \\ \cline{3-5}
		                                                                             &       & \multicolumn{1}{c|}{top-5} & \multicolumn{1}{c|}{top-10} & top-15 &       \\ \hline \hline
		Baseline                                                                     & 0.297 & \multicolumn{1}{c|}{0.571} & \multicolumn{1}{c|}{0.739}  & 0.850  & 0.544 \\ \hline
		\begin{tabular}[c]{@{}l@{}}Baseline + \\ Rectification\end{tabular}          &
		\textbf{0.540}                                                                &
		\multicolumn{1}{c|}{\textbf{0.622}}                                          &
		\multicolumn{1}{c|}{\textbf{0.808}}                                          &
		\textbf{0.940}                                                               &
		\textbf{0.545}                                                                                                                                                   \\ \hline
		State of the art                                                             & 0.300   & \multicolumn{1}{c|}{0.530}  & \multicolumn{1}{c|}{0.618}  & 0.650   & 0.544 \\ \hline
	\end{tabular}
\end{table}

\begin{table}[h!]
	\centering
	\caption{Inconsistency Results across Different Image Explaining Models}
	\footnotesize
	\label{tab2}
	\begin{tabular}{|l||c|ccc|c|}
		\hline
		\multirow{2}{*}{}                                                            &
		\multirow{2}{*}{\begin{tabular}[c]{@{}c@{}}Captioning \\ Model\tablefootnote{The Baseline models for Image Captioning is ~\cite{vinyals2015show} and the State of the art compared is ~\cite{li2022mplug}}\end{tabular}} &
		\multicolumn{3}{c|}{VQA Model\tablefootnote{The Baseline models for Visual Question Answering is ~\cite{yang2016stacked} and the State of the Art compared is ~\cite{kim2021vilt}}}                                              &
		\multirow{2}{*}{\begin{tabular}[c]{@{}c@{}}Prompt-based \\ Model {[}1{]}\end{tabular}}                                                                           \\ \cline{3-5}
		                                                                             &       & \multicolumn{1}{c|}{top-5} & \multicolumn{1}{c|}{top-10} & top-15 &       \\ \hline \hline
		Baseline                                                                     & 0.318 & \multicolumn{1}{c|}{0.627} & \multicolumn{1}{c|}{1.380}  & 2.130  & 0.096 \\ \hline
		\begin{tabular}[c]{@{}l@{}}Baseline + \\ Rectification\end{tabular}          &
		\textbf{0.20}                                                                &
		\multicolumn{1}{c|}{0.243}                                                   &
		\multicolumn{1}{c|}{\textbf{0.512}}                                          &
		\textbf{0.760}                                                               &
		\textbf{0.091}                                                                                                                                                   \\ \hline
		State of the art                                                             & 0.285   & \multicolumn{1}{c|}{\textbf{0.230}}  & \multicolumn{1}{c|}{0.551}  & 0.900   & 0.096 \\ \hline
	\end{tabular}
\end{table}

\subsection{Improvements over Image Captioning Models}
In the domain of Image Captioning, our architecture has shown remarkable improvements in generating more consistent captions having a higher extent of coverage compared to the baseline model with a minor reduction in BLEU score as shown in Table~\ref{tab1}. Utilizing MRCNN's object detection and segmentation capabilities as a rectifier, our model generated a completeness score of 0.54 which is an improvement of 81.81\% over that of the baseline model, as shown in Table~\ref{tab2} and an inconsistency reduction of 0.200 making it better than the baseline model by a 37.10\% , as shown in Table~\ref{tab3}. Furthermore, when implementing the same computation over the existing state-of-the-art model, we see that it has a lesser completeness score of 0.300 and a much higher inconsistency score of 0.285. These enhancements indicate that our model is able to give a more accurate description of the visual content.

\subsection{Improvements over Visual Question Answering}
For Visual Question Answering, the integration of MRCNN significantly augmented the accuracy and depth of responses. As shown in Table~\ref{tab2}, {\ir}   outperformed the baseline VQA across all the 3 configurations of k being 5, 10 and 15, with an improvement of 8.9\%, 9.33\% and 10.58\% for completeness respectively. Further, for the inconsistency, described in Table~\ref{tab3} the models surpass the baseline by achieving a 61\%, 62.73\% and 64.31\% drop in inconsistency score. In comparison with the State-of-the-art, our model is only outperformed by the former when k is set to 5 with a score of 0.130 higher inconsistency. However, for k as 10 and 15, ~\ir~ is able to generate a reduced inconsistency of 7.1\% and 15\% respectively. This indicates a more comprehensive explanation of the visual content, enabling the model to provide answers that are both precise and contextually rich.

\subsection{Improvements over Prompt Based AI with GPT-4}
Perhaps most notably, our proposed architecture has enhanced the capabilities of GPT-4 in the domain of conversational or Prompt Based AI. By rectifying the inputs with detailed visual analyses from MRCNN, the model demonstrated a subtle enhancement in generating responses that are contextually aware and grounded in the visual reality presented to it. The integration led to a 5.2\% improvement in the inconsistency and an improvement in completeness of 0.02\% over GPT-4 operating without the rectification layer. This advancement underscores the potential of our architecture to refine generative processes of Prompt Based AI, even when applied to State-of-the-art models such as GPT-4.

\subsection{Prominence based Completeness}
Further, in our analysis of \ir's performance, we conduct a preliminary analysis on the correlation between completeness and meaningfulness of the objects with respect to their size and prominence in the image. In order to conduct this, we design a "Prominence-based Completeness" filter, which strategically excluded objects less than 5\%, 10\%, and 15\% of the entire image's area. This gives very interesting insight to \ir where we compare the completeness between the detection capability of VQA and the quality of caption of the underlying image captioning model as explained in the Table~\ref{PROMINENCE}. For the image captioning domain, we notice that when the background objects are removed, leaving behind just foreground objects that are crucial for the caption, there is an increase in completeness. However, when further increasing the filter, we start missing out the crucial objects as well, resulting in a dip in the completeness. On contraire, the VQA domain gives an initial drop in the completeness followed by a subtle rise. This behaviour is a result of the object detection approach of the VQA model which is not affected by any inter-object relation and is able to recognize more salient objects with the increase in filter ratio. These enhancements were most pronounced at the 85\% threshold with \(k=10\), where the balance between detail retention and noise reduction jumped a little higher than the counterparts. The results of this preliminary analysis shows the potential of using image captioning modules in a progressing manner to get much finer explanations.

\begin{table}[]
\centering
\caption{``Prominence-based Completeness'' filter for different values of filter with k=95,90,85, which removes any objects accounting area less than 5\%,10\%,15\% respectively, of the entire image}
\label{PROMINENCE}
\begin{tabular}{|c|c|ccc|}
\hline
\multicolumn{1}{|c|}{\multirow{2}{*}{\textbf{\begin{tabular}[c]{@{}c@{}}Prominence\\ Ratio\end{tabular}}}} &
  \multicolumn{1}{c|}{\multirow{2}{*}{\textbf{\begin{tabular}[c]{@{}c@{}}Image\\ Captioning\end{tabular}}}} &
  \multicolumn{3}{c|}{\textbf{VQA (topk)}} \\ \cline{3-5} 
\multicolumn{1}{|c|}{} &
  \multicolumn{1}{c|}{} &
  \multicolumn{1}{c|}{\textbf{top-5}} &
  \multicolumn{1}{c|}{\textbf{top-10}} &
  \multicolumn{1}{c|}{\textbf{top-15}} \\ \hline \hline
\textbf{100\%} & 0.54 & \multicolumn{1}{c|}{0.62} & \multicolumn{1}{c|}{0.80} & 0.94 \\ \hline
\textbf{95\%}  & 0.83 & \multicolumn{1}{c|}{0.35} & \multicolumn{1}{c|}{0.46} & 0.54 \\ \hline
\textbf{90\%}  & 0.70 & \multicolumn{1}{c|}{0.36} & \multicolumn{1}{c|}{0.47} & 0.55 \\ \hline
\textbf{85\%}  & 0.58 & \multicolumn{1}{c|}{0.37} & \multicolumn{1}{c|}{0.55} & 0.56 \\ \hline
\end{tabular}
\end{table}



The results showcases the efficacy of ~\ir~ in enhancing the interpretability and accuracy of AI models in understanding and explaining images. By leveraging the precise object detection and segmentation capabilities of MRCNN, our model not only improves the quality of image-based explanations but also enriches the interaction between AI and users in conversational contexts. Notably, the implementation of prominence-based filtering has markedly enhanced the clarity and relevance of the explanations, contributing significantly to the observed improvements. The significant improvements observed across all domains highlight the potential of our architecture to set a new benchmark in the fields of image captioning, visual question answering, and Prompt Based AI.


\section{Conclusion} \label{sec:conclusion}
Image explanation is an interesting problem in Deep Learning models. In particular, the existing architectures for Image Captioning, VQA or prompt-based engines suffers from hallucinating objects and/or missing the presence of objects. In this article, we present rectification-based framework (called \ir), built atop existing image explaining models, which inspects over multiple options for the output, harnesses the precision of Mask R-CNN as a rectifier, and refines the output of existing explainable models. We showcase the improvement of our proposed framework in identifying the image objects in a more consistent and complete manner. Further, our rectification framework is generic enough to be plugged atop any image explaining model to make it more robust. Since no work in this field is perfect and for a novel architecture like this, proposing a unique approach for rectification, there is a lot of room for improvement. One of the drawbacks is the dependency of the baseline models and the rectifier to be trained on the same dataset for an accurate rectification. The reliance of both the baseline image explaining model and the rectifier limits the rectification ability of our framework. For our future work, we plan to propose an enhanced version of the proposed metrics, which incorporates a weight factor of the rectified objects. Further we intend to integrate our proposed rectification layer with other forms of multimodal domains and aim to ensemble the outcomes from various image explaining architectures to further provide a comprehensive and robust explanation. As we continue to refine and expand the scope of our work, we remain committed to the rectification approach of our work and aim to make much-enhanced version of the same.

\bibliographystyle{ACM-Reference-Format}
\bibliography{Reference}

\end{document}